\documentclass{llncs}
\usepackage{amsmath,graphicx,booktabs}
\usepackage{epstopdf,url}
\usepackage{makeidx,amssymb}  
\usepackage{caption}
\usepackage[linesnumbered,ruled,vlined]{algorithm2e}
\usepackage{array}
\newcolumntype{C}[1]{>{\centering\arraybackslash}p{#1}}

\begin{document}

\title{
Iterative Multi-domain Regularized Deep Learning for Anatomical Structure Detection and Segmentation from Ultrasound Images}
%
%

\author{
Hao Chen\inst{1,2} 
\and Yefeng Zheng\inst{2}
\and Jin-Hyeong Park\inst{2}
\and Pheng-Ann Heng\inst{1}
\and \\ S. Kevin Zhou\inst{2}
}


\institute{
Dept. of Computer Science and Engineering, The Chinese University of Hong Kong
\and Medical Imaging Technologies, Siemens Healthcare, Princeton, NJ, USA
}

%

\maketitle

\begin{abstract}

Accurate detection and segmentation of anatomical structures from ultrasound images are crucial for clinical diagnosis and biometric measurements.
Although ultrasound imaging has been widely used with superiorities such as low cost and portability, the fuzzy border definition and existence of abounding artifacts pose great challenges for automatically detecting and segmenting the complex anatomical structures.
In this paper, we propose a multi-domain regularized deep learning method to address this challenging problem.
By leveraging the transfer learning from cross domains, the feature representations are effectively enhanced.
The results are further improved by the iterative refinement.
Moreover, our method is quite efficient by taking advantage of a fully convolutional network, which is formulated as an end-to-end learning framework of detection and segmentation.
Extensive experimental results on a large-scale database corroborated that our method achieved a superior detection and segmentation accuracy, outperforming other methods by a significant margin and demonstrating competitive capability even compared to human performance.

\end{abstract}
\section{Introduction}
\label{sec:intro}
\begin{figure}[t]
\centering
  \includegraphics[width=.9\linewidth]{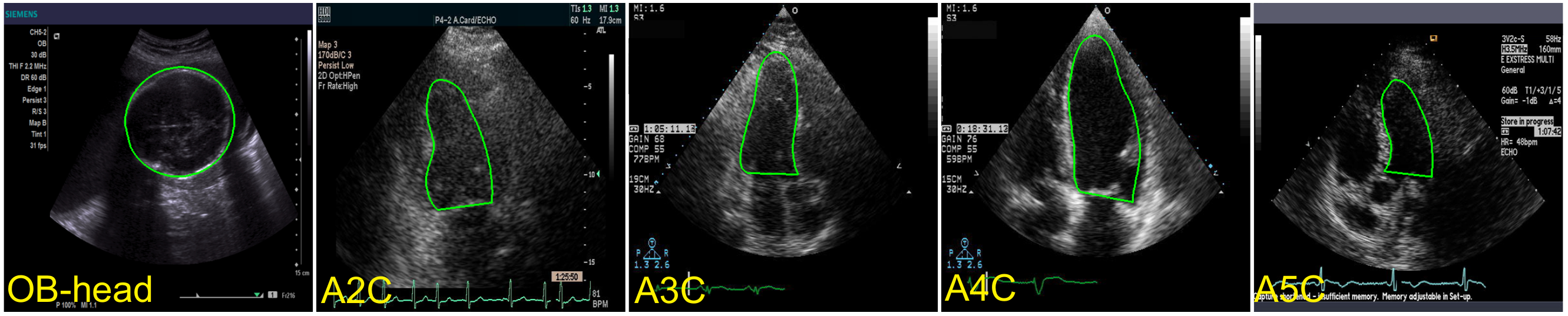}
\caption{Typical examples of anatomical structure detection and segmentation in different views (expert annotations are overlaid on the images with green contours). }
\label{fig:example}
\vspace{-10pt}
\end{figure}

The ultrasound imaging has been a powerful tool for visualizing the complex anatomical structures due to its superior advantages, e.g., portability, real-time imaging, low-cost and free of radiation.
From the ultrasound images, clinically relevant measurements are often derived for clinical diagnosis, based on the reliable detection and segmentation of anatomical structures, such as the left ventricle (LV) on apical-2-chamber (A2C), A3C, A4C and A5C views~\cite{zhou2007shape} and circumference measurement of the head in an obstetric exam (OB-head), etc.
Typical examples with expert annotations are illustrated in Fig.~\ref{fig:example}.
The routine delineation by manual power suffers from several issues such as tedious efforts, time-consuming work and limited reproducibility.
Therefore, automated methods are highly demanded to reduce the workload on clinicians and improve the routine efficiency as well as reliability.
However, detection and segmentation of anatomical structures from ultrasound images remain a challenging task for several reasons.
First, the overfitting issue can lead to inferior performance for supervised learning based methods when dealing with a limited number of training datasets, which is common for most medical imaging computing problems.
Second, low image quality such as fuzzy border and abounding artifacts (e.g., acoustic shadows) poses great challenges for automated methods.
Finally, the image characteristics may differ significantly depending on the ultrasound machines and operator settings.

Considerable progress has been achieved on this challenging problem over the past few years.
Previous studies usually consist of individual steps with piece-wise design for detection and segmentation separately~\cite{georgescu2005database,zhou2012discriminative,zhou2007shape}.
Georgescu et al.~\cite{georgescu2005database} proposed a database-guided method for segmenting anatomical structures with complex appearance. Although preliminary good performance has been evidenced, the detection robustness may be bounded by the low-level handcrafted features, i.e., Haar-like features.
Compared with handcrafted features, there are more evidences proving that high-level feature representations learned from deep neural networks can usually achieve better performance.
An approach using deep belief networks (DBN) and derivative-based search strategy was presented in~\cite{carneiro2012segmentation}.
Recently, convolutional neural networks (CNNs) have gained prevalence on image classification and segmentation tasks~\cite{long2015fully}. 

In order to tackle the aforementioned challenges, we present a multi-domain regularized deep learning method to reduce the overfitting issue due to limited medical dataset. 
With elegantly designed CNN architectures, our method can harness the knowledge across the large amounts of cross-domain data for effective generic feature representations.
Specifically, we propose a unified framework that leverages fully convolutional networks~\cite{long2015fully} for end-to-end learning and inference; hence the detection and segmentation efficiency can be greatly boosted.
Extensive experimental results on a large-scale database demonstrated that our method incorporating multi-domain regularization with iterative refinement can achieve a superior segmentation accuracy, surpassing other methods by a significant margin on various evaluation metrics.


\section{Method}
\label{sec:method}
\subsection{Semantic Segmentation with Fully Convolutional Networks}
Recently, fully convolutional networks (FCN), i.e, a variant of CNN, achieved the state-of-the-art performance on image segmentation related tasks~\cite{long2015fully}. 
Such a great success is mostly attributed to the outstanding capability in feature representations for dense classifications.
The whole network can be trained in an end-to-end (image-to-image) fashion, which takes an image as input and outputs the probability map directly. 
The architecture basically contains two modules including a downsampling path and an upsampling path.
The downsampling path contains convolutional and max-pooling layers, which are extensively used in the convolutional neural networks for image classification tasks. 
The upsampling path contains deconvolutional layers (a.k.a. backwards strided convolution~\cite{long2015fully}), which upsample the feature maps and output the score masks.

\subsection{Multi-domain Regularized FCN}
Despite that considerable success has been achieved on image segmentation tasks in computer vision, the inherent issues in ultrasound imaging such as speckle noise and acoustic shadows pose difficulties for segmentation related tasks~\cite{carneiro2012segmentation,georgescu2005database,zhou2012discriminative,zhou2007shape}.
Furthermore, the limited availability of annotated medical datasets further deteriorates the situation for supervised learning based methods.
Previous studies have evidenced the efficacy of transfer learning from natural image domain to medical image domain (particularly the ultrasound image domain) on image classification tasks~\cite{chen2015standard}. 
However, to the best of our knowledge, how to leverage the ultrasound images from different domains for learning effective feature representations on the task of anatomical structure detection and segmentation has not been fully explored.
\begin{figure}[t]
\centering
  \includegraphics[width=.85\linewidth]{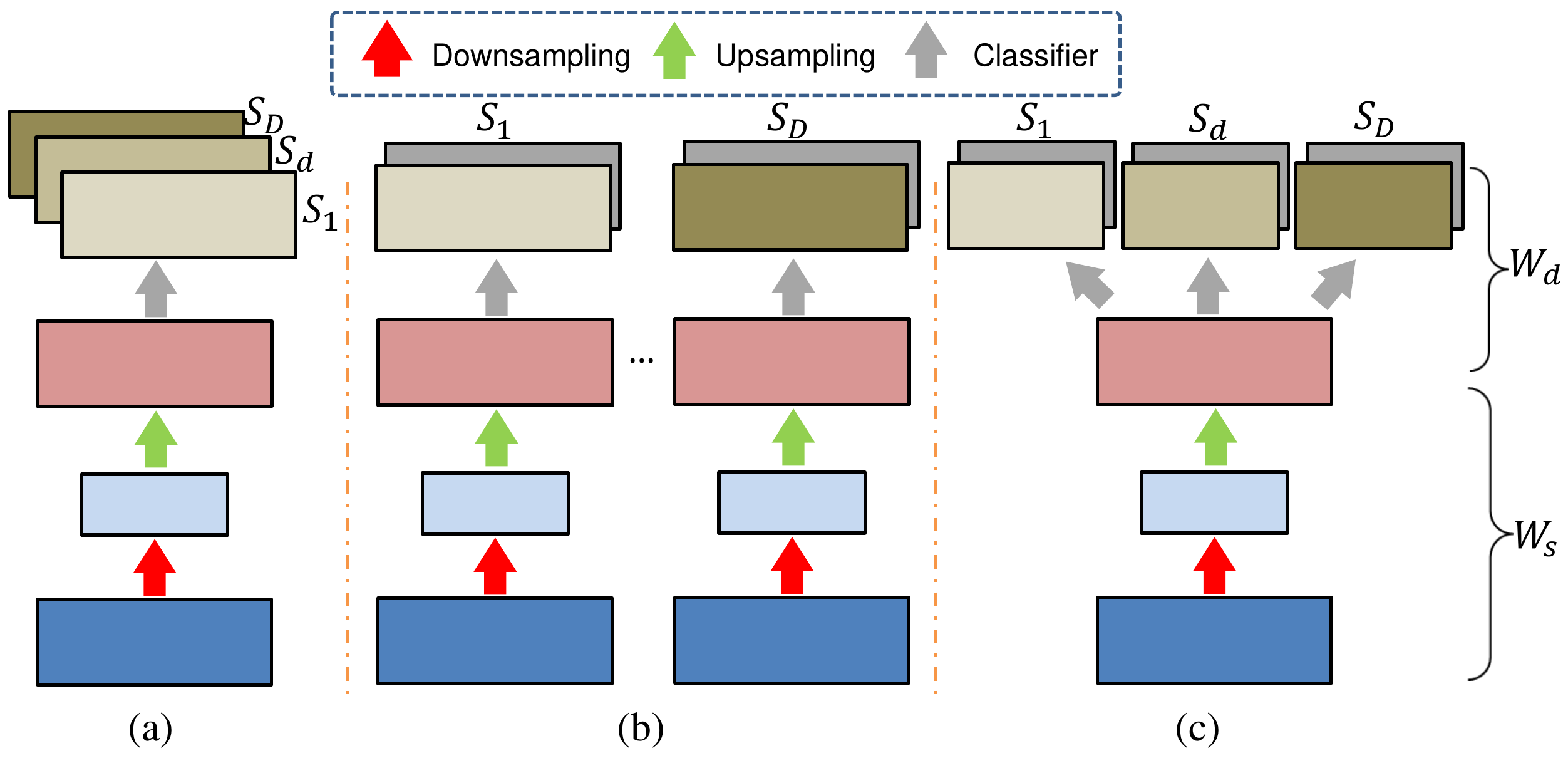}
\caption{The overview of different network architectures: (a) multi-label FCN (ML-FCN), (b) single-domain FCN (SD-FCN), (c) multi-domain regularized FCN (MD-FCN). }
\label{fig:framework}
\end{figure}

In this section, we conduct extensive studies on how to elaborately make full use of cross-domain data by leveraging transfer learning in deep neural networks.
Here we refer each ultrasound view as one domain, in which a different anatomical structure or a different 2D view of the same structure is presented.
We compare three different FCN architectures as shown in Fig.~\ref{fig:framework}.
The first framework of multi-label FCN (ML-FCN) follows the typical structure of FCN, which takes an image as input and directly outputs the segmentation masks of several domains.
The second one, i.e., single-domain FCN (SD-FCN), is trained by minimizing the pixel-wise cross-entropy loss on the ultrasound images of a single domain.
Although the ML-FCN uses all data from cross-domain, it mostly learns the same representation for all domains, which lacks the domain-specific discrimination, while the SD-FCN is the other way around. 
In order to address the potential issue of insufficient training data in one single domain, we present a multi-domain regularized FCN (MD-FCN) to further improve the performance.
This is inspired by the observation that feature representations in deep neural networks transit from generality to specificity from lower to higher layers in the hierarchical network~\cite{yosinski2014transferable}.
Thus, the lower layers should be exploited among different domains while the higher layers are enhanced with discrimination capability of a single domain.
Therefore, we design an architecture consisting of~\emph{domain-generic} layers and~\emph{domain-specific} layers as shown in Fig.~\ref{fig:framework} (c), which can be jointly trained on ultrasound images from different domains.
The weights $W_s$ of generic layers are shared and updated in all domains for strengthening the feature representations in low-level cues, while the weights $W_d$ of specific layers are learned in each domain for discriminating the background and anatomical structures based on high-level semantic information.
A softmax classification layer is exploited to generate the probability map.
Finally, the MD-FCN is trained via a mini-batch gradient descent method by minimizing following loss function: 
\begin{gather}\label{multi-task1}
  \mathcal{L} =  -\sum_{d=1}^{D} \sum_{n=1}^{N_d} \sum_{x \in I_{n}^{d}} \log p(y_x|x;W_s,W_d) + \frac{\lambda}{2} \sum_{d=1}^{D} \sum_{n=1}^{N_d} (||W_d||_2^{2}+||W_s||_2^2)   
\end{gather}
where the first part is the fidelity term defined with cross-entropy loss and the second one is the regularization term (weight $\lambda$ controls the balance). 
Specifically, $D$ is the total number of domains; $N_d$ is the number of training images in the domain indexed by $d$, $p(y_x|x;W_s,W_d)$ denotes the predicted probability regarding true label $y_x$ for pixel $x$ in the $n$th image $I_{n}^{d}$ from the $d$th domain.

\subsection{Refinement with Iterative MD-FCN}
There are two potential problems in the FCN based methods for ultrasound image segmentation: 1) the down-sampling path reduces the resolution of input, which causes inaccurate localization due to the information loss through the network; and 2) the existence of confounding structures and acoustic shadows can cause false positive results within the non-anatomical structural regions.
In order to tackle these problems, we develop an iterative version of MD-FCN (iMD-FCN) for segmentation refinement.
Specifically, we gradually select the input region only containing the anatomical structures according to an interleaved segmentation-detection strategy, thus the network can concentrate on segmenting the anatomical structures after the irrelevant information is removed.
We crop the region enclosing detected anatomical structures (with around 20\% background context regions included), which are detected by the MD-FCN based on the whole input image, and then upsample the region to a larger image (e.g., $480\times480$ pixels in our experiments) for segmentation.
This attention mechanism can effectively remove confounding background regions and reduce the information loss to some extent.
With this iterative step, more accurate results can be generated with boundary refined. 
The iterative process is terminated if the Dice ratio of consecutive segmentation results stops changing.
For training the iMD-FCN model, training samples are generated by cropping the rectangular regions containing the anatomical structures (with 10-50\% background regions) from the training data given the expert annotations.

\section{Experiments}
\label{sec:exp}
\subsection{Datasets and Pre-processing}
To validate the effectiveness of our method under the setting of different ultrasound machines, the images were acquired by ultrasound scanners from multiple vendors including Philips (ATL) and Siemens (Sequoia and SC2000) in our experiments.
We conducted extensive experiments on five ultrasound views including OB-head, A2C, A3C, A4C and A5C.
The summary of datasets can be found in Table~\ref{table:details}.
To the best of our knowledge, this is the largest ultrasound database evaluated on the task of anatomical structure detection and segmentation so far.
The anatomical structures were annotated by two clinical experts with more than five-year experience in ultrasound examination (one is regarded as the ground truth and the other one is used for human performance evaluation).

\begin{table}[t]
\caption{The number of images in our ultrasound dataset.}
\small
\label{table:details}
\begin{center}
\begin{tabular}{lC{8ex}C{8ex}C{8ex}C{8ex}C{8ex}C{8ex}C{8ex}}
\toprule[1pt]
Dataset     & OB & A2C & A3C & A4C & A5C  & Total\\
\midrule
Training     &1,313  &9,315 & 8,228 & 12,500 & 3,005 &  34,361\\
Test     &329  &2,314 & 2,055 & 3,118 & 717 & 8,533 \\
\bottomrule
\end{tabular}
\end{center}
\end{table}
\subsection{Implementation Details}
In our experiments, we empirically found that deep neural networks with a random initialization took a long time to converge and achieved inferior performance on this challenging task.
Previous studies have indicated that CNN leveraging transfer learning from large-scale natural image dataset can achieve much better performance~\cite{chen2015standard,tajbakhsh2016convolutional}. 
Therefore, we utilized an off-the-shelf DeepLab model~\cite{chen2014semantic} (trained on the PASCAL VOC dataset) for initializing the downsampling layers of our FCN model.
We then added upsampling layers and discriminant layers for segmenting anatomical structures from ultrasound images.
We implemented our method based on the Caffe library~\cite{jia2014caffe} on a workstation with one GTX Titan X GPU.
It took about 6 hours to train the iMD-FCN models and less than 0.5 seconds to process one test image with size $480\times 480$ pixels.

\begin{figure}[t]
\centering
  \includegraphics[width=1.\linewidth]{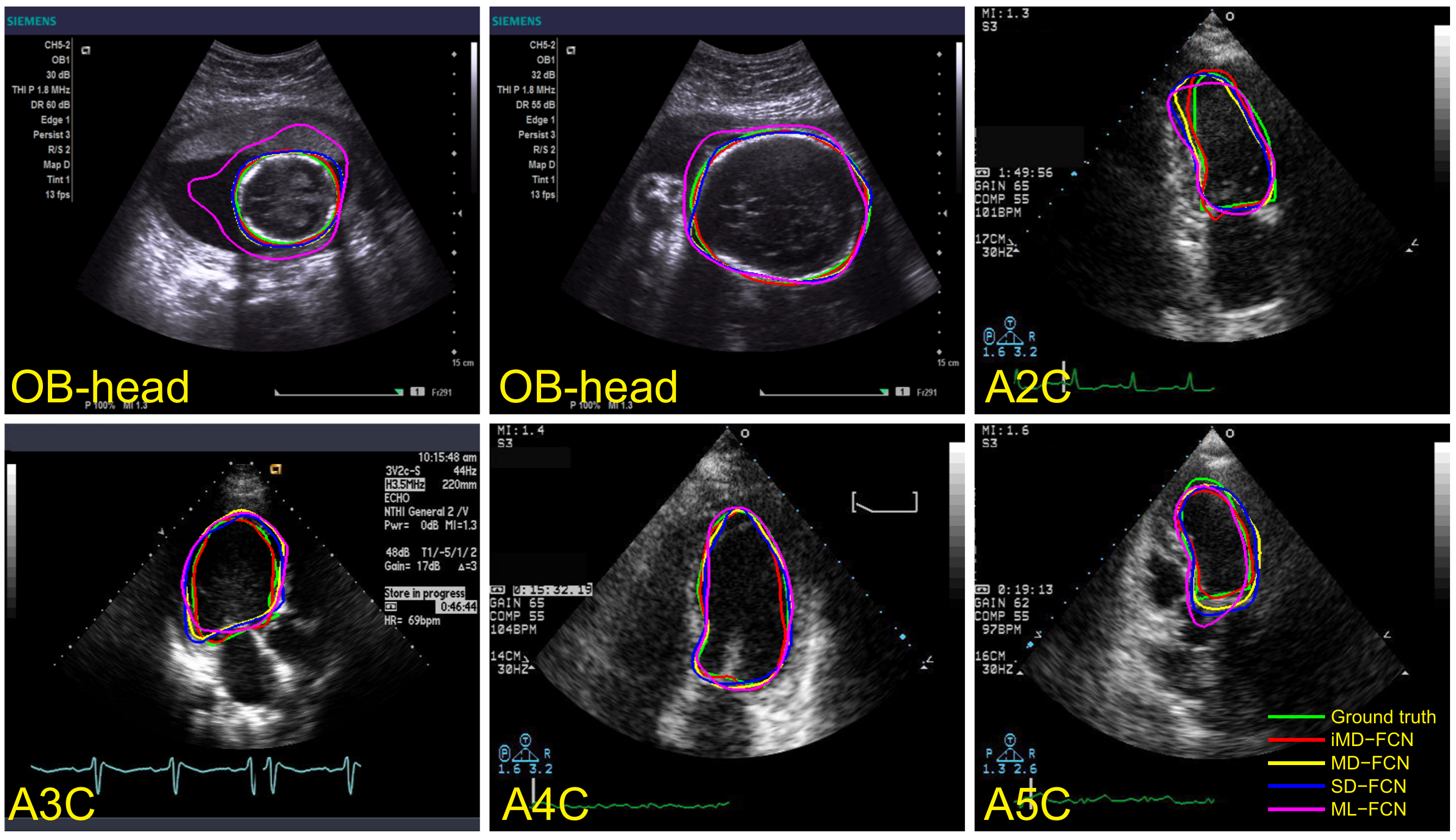}
\caption{Examples of detection and segmentation results from different methods (results of different methods are overlaid on the images with different colors). }
\label{fig:result1}
\end{figure}

\begin{table}[t]
    \resizebox{1.0\linewidth}{!}{
    \begin{minipage}{.5\linewidth}
    \caption{F1 score of detection results.}
    \small
      \label{table:det_result}
      \centering
\begin{tabular}{lccccc}
\toprule[1pt]
Method      & OB & A2C & A3C & A4C & A5C \\
\midrule
ML-FCN &0.972  &0.974 & 0.928 & 0.952 & 0.884 \\
SD-FCN &0.999  &0.989 & 0.969 & 0.978 & 0.904 \\
MD-FCN &0.999  &0.990 & 0.973 & 0.979 & 0.979 \\
iMD-FCN &\bf{1.0}  & \bf{0.995}  & \bf{0.996}  & \bf{0.993}  & \bf{0.996}  \\
DGS~\cite{georgescu2005database}  &-  &0.986 & - & 0.992 & - \\
\midrule
Human  &1.0  &1.0 & 1.0 & 1.0 & 1.0 \\
\bottomrule
\end{tabular}
    \end{minipage}%
    \hspace{1ex}
    \begin{minipage}{.5\linewidth}
      \centering
\caption{Dice ratio of segmentation results.}
\small
\label{table:seg_result}
\begin{tabular}{lccccc}
\toprule[1pt]
Method      & OB   & A2C & A3C & A4C & A5C \\
\midrule
ML-FCN  &0.860  &0.826 & 0.773 & 0.804 & 0.738  \\
SD-FCN &0.942  &0.851 & 0.811 & 0.833 & 0.768 \\
MD-FCN  &0.950  & 0.865 & 0.822  & 0.837 & 0.797 \\
iMD-FCN  &\bf{0.961}  & \bf{0.875} & \bf{0.864}  & \bf{0.879} & \bf{0.891} \\ 
DGS~\cite{georgescu2005database}  &-  &0.832 & - &0.844 & -\\
\midrule
Human  &0.971  &0.908 & 0.858 & 0.917 & 0.909\\
\bottomrule
\end{tabular}
    \end{minipage}
}
\end{table}

\subsection{Evaluation Metrics}
For a comprehensive evaluation, the performance on detection, segmentation and shape similarity of different methods were compared, respectively.
\\
\textbf{Detection.}
For the detection evaluation, the metric F1 score is utilized, which is the harmonic mean of precision and recall. 
A segmented object is regarded as a true positive if the Jaccard index between the segmented object and the ground truth is equal to or larger than 0.5, otherwise it's considered as a false positive.
The ground truth is considered as a false negative if the Jaccard index between the ground truth and the segmented object is less than 0.5.
\\
\textbf{Segmentation.}
The Dice ratio is employed for the segmentation evaluation. 
\\
\textbf{Shape Similarity.}
The shape similarity is measured by using the Hausdorff distance between the shape of segmented object and that of the ground truth. 
If no segmented object is generated, we take the whole image as segmentation and calculate the Hausdorff distance accordingly.

\begin{table}[t]
\caption{Hausdorff distance of shape similarity results (unit: pixels).}
\small
\label{table:shape_result}
\begin{center}
\begin{tabular}{lccccc}
\toprule[1pt]
Method      & OB   & A2C & A3C & A4C & A5C \\
\midrule
ML-FCN &30.57  & 25.36 & 29.95 & 27.75  & 30.61 \\
SD-FCN &14.93  &23.76 & 24.69 & 25.15 & 30.47 \\
MD-FCN  &11.73  & 20.38 & 23.69  & 24.03 & 23.73 \\
iMD-FCN   &\bf{9.21}  & \bf{18.76} & \bf{19.70}  & \bf{18.22} & \bf{16.31} \\
DGS~\cite{georgescu2005database}  &-  & 20.80 & - & 18.87 & -\\
\midrule
Human  &5.19  & 14.32 & 17.60 & 11.52 & 12.78\\
\bottomrule
\end{tabular}
\end{center}
\end{table}
\subsection{Evaluation and Comparison}
We compared results of different architectures in Fig.~\ref{fig:framework} and one state-of-the-art method utilizing database-guided segmentation, referred as DGS~\cite{georgescu2005database}.
In order to demonstrate the difficulty of confronting problem, we also compared the performance of two human experts. 
The detection results of all methods are shown in Table~\ref{table:det_result}.
The iMD-FCN method achieved the best performance on all views with the highest F1 score, indicating that it can detect the anatomical structures most robustly.
The averaged segmentation and shape similarity results on all images are shown in Table~\ref{table:seg_result} and Table~\ref{table:shape_result}, respectively.
The method of ML-FCN achieved inferior results compared to other methods, due to the existence of confounding anatomical structures such as in A5C that the architecture cannot handle them properly.
Furthermore, the MD-FCN method achieved marginal improvements than SD-FCN method in the domain where training data is abundant such as A4C, while the improvement is significant in the domains with limited training data such as A5C (Hausdorff distance decreased from 30.47 to 23.73 pixels).
This highlighted the efficacy of incorporating multi-domain regularization in learning powerful feature representations, which is quite important in the situation when training datasets are limited.
The iMD-FCN method outperformed all the other methods by a significant margin and the success was attributed to the combination of cross-domain transfer learning and iterative segmentation refinement.
In addition, it surpassed human performance on the segmentation of A3C and was comparable on the views of OB and A5C, which verified its effectiveness clearly.

Typical examples of detected and segmented anatomical structures are illustrated in Fig.~\ref{fig:result1}.
We can see that MD-FCN can achieve more accurate results qualitatively than SD-FCN, as the former one learned more powerful feature representations by leveraging the knowledge learned on the large datasets from cross domains.
It is notable that our method iMD-FCN can detect and segment the anatomical structures despite the large variations of anatomical structures (such as the large scale variation in OB-head).
It is also robust to the inferior image quality, even with artifacts such as shadows and speckle noise (as shown in the A2C example).
Furthermore, our method can segment the anatomical structures accurately in the situation where the boundary is not clear (see A2C and A4C examples), which is even challenging for experts.

\section{Conclusion}
In this paper, we proposed an effective iterative multi-domain regularized deep learning method for anatomical structure detection and segmentation from ultrasound images.
We anticipate that multi-domain regularization in deep learning will become an important method for the medical imaging computing community, which faces the challenge of insufficient training data.
In the future work, we will incorporate advanced shape regression methods to further improve the segmentation performance. 
\\
\\
\textbf{Acknowledgements.}
The technical work was done when Hao Chen worked for Siemens Healthcare as an intern. We thank Research Grants Council of the Hong Kong Special Administrative Region (No. CUHK 412513) and Hong Kong-Shenzhen Innovation Circle Funding Program (No. GHP/002/13SZ and SGLH20131010151755080) for supporting conference attendance and others.

\bibliographystyle{splncs03}
\bibliography{refs}
\end{document}